\documentclass[runningheads]{llncs}
\usepackage{graphicx}
\usepackage{amsmath,amssymb}
\usepackage{color}
\usepackage[width=122mm,left=12mm,paperwidth=146mm,height=193mm,top=12mm,paperheight=217mm]{geometry}
\usepackage{float}
\begin{document}
\pagestyle{headings}
\mainmatter
\def\ECCV16SubNumber{***}  % Insert your submission number here

\title{Early Glaucoma Detection using Deep Learning with Multiple Datasets of Fundus Images}

% \titlerunning{ECCV-16 submission ID \ECCV16SubNumber}
% \authorrunning{ECCV-16 submission ID \ECCV16SubNumber}

\author{Rishiraj Paul Chowdhury and Nirmit Shekar Karkera}
\institute{Course CSCI 5922, University of Colorado Boulder}

\maketitle

\begin{abstract}
Glaucoma is an eye condition that damages the optic nerve, which can lead to vision loss or blindness. This condition affects individuals worldwide, but early glaucoma detection can help diagnose the condition faster and enhance patient treatment. Traditional diagnostic methods, such as Tonometry, Ophthalmoscopy, and Gonioscopy are costly, invasive to the eye, and require a medical specialist. However, non-invasive methods such as deep-learning approaches based on fundus images of the eye show promising results but such architectures are  typically trained on single datasets, which limits their practical generalizability to different patients. In this project, we develop a convolutional neural network (CNN) model based on the EfficientNet architecture, trained sequentially across the ACRIMA, ORIGA, and RIM-ONE datasets of fundus images, to enhance diagnostic accuracy and model generalizability. By conducting experiments on the trained model and evaluating metrics such as accuracy, sensitivity, specificity, and AUC-ROC, we demonstrate this method’s capability for improved glaucoma detection and its potential use in clinical data for early detection. Ultimately, our work aims to deliver an accurate, easy-to-use, and scalable model for non-invasive early glaucoma screening, which contributes to better patient treatment through timely clinical intervention.

\end{abstract}
\section{Introduction}

% Paragraph 1: Explain the motivation for your work; e.g., Why anyone should care? What are the desired benefits?
Glaucoma, an eye-disease marked by the gradual deterioration of the optic nerve, has been recognized as one of the primary contributors to permanent vision loss worldwide. When identified at an early stage and addressed promptly with appropriate care, the likelihood of serious sight degradation can be greatly minimized. However, this condition is frequently left undiagnosed until noticeable damage to vision has already taken place . Present-day diagnostic techniques for glaucoma, including eye pressure assessments and optical coherence imaging, depend upon advanced machinery, professional knowledge, and considerable resources—posing difficulties for broad and consistent screening, especially in medically neglected areas. The creation of reliable, flexible, and easily deployable diagnostic systems through algorithmic techniques has the potential to close this pressing healthcare divide and deliver valuable public health improvements by reducing preventable blindness through early-stage identification.

% Paragraph 2: Explain why existing solutions are inadequate for the motivated problem; e.g., Is there a gap in what exists? Is there a weakness in existing approaches?
Encouraging results in the computerized detection of glaucoma have been revealed through continuous advancements in artificial intelligence strategies, particularly by employing convolutional neural architectures (CNNs) with retinal fundus photographs. Nonetheless, the majority of contemporary algorithms are constructed and assessed on isolated, uniform datasets, resulting in diminished accuracy when transferred to actual medical settings due to variations in image-capturing environments, patient characteristics, and data-specific distortions. These drawbacks considerably constrain the medical usability and adaptability of prevailing methodologies, indicating substantial potential for enhancement in reliability and performance.

% Paragraph 3: Explain what you are proposing, what is new about your idea, and why you believe this solution will be better than previous solutions; e.g., Are you asking a new question, establishing a new methodology to solve a problem, building a new software tool, or offering greater understanding about existing methods/tools?
Within this study, the development and training of a resilient convolutional neural network (CNN)-driven deep learning framework have been undertaken using several varied glaucoma detection datasets. Specifically, the ACRIMA, ORIGA, and RIM-ONE collections of retinal fundus images have been utilized. The primary advancement introduced in our approach lies in the structured combination and stepwise adaptation of the model throughout these heterogeneous datasets, aiming to significantly enhance the model’s diagnostic accuracy and generalizability. Unlike previous studies that rely predominantly on single-dataset training and testing, our approach explicitly addresses variations among data collections, thereby enhancing the model’s dependability for effective, real-life implementation. This research aims to establish a new methodological standard in glaucoma detection through a generalized deep learning framework capable of accurately identifying glaucoma across varied imaging contexts and patient populations.
\section{Related Work}
\paragraph{\textbf{Deep learning techniques for glaucoma detection}}
Fu et al. ~\cite{fu2018disc} in Disc-aware ensemble network for glaucoma screening from fundus image  proposed a disc-aware  network in the paper that used both global and local features from fundus images for glaucoma classification and achieved a strong performance on the benchmark datasets. Again, Christopher et al.~\cite{christopher2020performance} in Performance of deep learning architectures for detection of glaucoma using fundus photographs performed an evaluation of several CNN-based architectures for glaucoma detection and reported high classification accuracy when using color fundus images. Our proposed work differs from these research studies because, although they demonstrate the effectiveness of CNN classifiers on fundus images but they typically rely on a single dataset for training and evaluation. This limitation hinders the ability of such models to generalize well across diverse image sets, imaging equipment, and acquisition strategies. Thus, it is addressed by our project by designing a sequential training and fine-tuning process on three publicly available glaucoma image datasets: ACRIMA, ORIGA, and RIM-ONE, which aims to reduce dataset-specific bias, improve generalization, and perform better for real-world fundus images.

\paragraph{\textbf{Fundus image preprocessing strategies}}
Mookiah et al.~\cite{mookiah2012automated} in Automated diagnosis of glaucoma using fundus images: A review provides a comprehensive review of preprocessing techniques used for automated glaucoma detection, highlighting the use of optic disc segmentation, vessel extraction, and contrast enhancement to improve classification accuracy. Orlando et al.~\cite{orlando2017discriminatively} in A discriminatively trained fully connected conditional random field model for blood vessel segmentation in fundus images developed a conditional random field model trained to segment retinal blood vessels, demonstrating the value of detailed structural features in downstream tasks like disease classification. Our work is different from these papers because they either focus on extensive segmentation-based preprocessing (e.g., optic disc and cup segmentation, vessel detection) or apply traditional image enhancement techniques such as contrast adjustment or morphological filtering. While these methods have proven valuable, they are often computationally intensive and can introduce dependencies on intermediate models. In contrast, our approach adopts lightweight, general-purpose preprocessing steps—such as resizing, histogram equalization, and color normalization—designed to unify the image quality across datasets without requiring any segmentation or vessel extraction. This allows the CNN model to learn directly from raw image content and promotes scalability across datasets and imaging conditions.

\paragraph{\textbf{Computer Vision Techniques in Medical Applications}}
Ajitha et al.~\cite{ajitha2021glaucoma} in Identification of glaucoma from fundus images using deep learning techniques made a 13-layer CNN model for detecting glaucoma using fundus images collected from different datasets, such as HRF, Origa, Drishti-GS1, and a local hospital. Their model got an accuracy of 95.61 percent when used with an SVM classifier, demonstrating strong accuracy and possible clinical importance. Similarly, Shoukat et al.~\cite{shoukat2023diagnosis} in Automatic diagnosis of glaucoma from retinal images using deep learning approach introduced an automated glaucoma detection system based on the ResNet-50 architecture, utilizing grayscale retinal images obtained from four publicly available datasets—G1020, DRISHTI-GS, RIM-ONE, and ORIGA. 

In contrast, our study introduces several distinct contributions. Whereas previous models often relied on a single dataset or merged datasets without addressing variability between them, we implement a novel sequential fine-tuning approach across three diverse datasets—ACRIMA, ORIGA, and RIM-ONE—end-eavoring to enhance generalization across varying image conditions and patient groups. we will use the EfficientNet-B0 architecture, which provides a good trade-off between accuracy and computational cost than the CNN and ResNet-based models used in previous work. Instead of relying on segmentation-heavy preprocessing, our pipeline uses a streamlined method that includes resizing, histogram equalization, and color normalization—enabling efficient and scalable, end-to-end learning. To assess the robustness of our model, we conduct targeted experiments, including evaluations on unseen datasets. On top of that, we guaranteed reproducibility by providing an open-source, adaptive framework, setting up our method as a feasible and adjustable solution for real-world glaucoma testing scenarios.
\section{Methodology}

To detect early signs of glaucoma using retinal fundus images, we design a deep learning pipeline trained on three publicly available datasets — ACRIMA, ORIGA, and RIM-ONE. This pipeline has four parts i.e. preprocessing of fundus images, designing and modifying the architecture, training and fine-tuning the model across datasets, and finally evaluating the performance on different metrics.

\subsection{Preprocessing of Multi-Datasets}
We start by standardizing the fundus images coming from different datasets to obtain consistency across it. All images are resized to 224$\times$224 pixels and converted to RGB format that match the input dimensions of our neural networks model. We also perform normalization on the image sets for a stable input distribution and faster model training. Improving visual uniformity, we perform histogram equalization, contrast normalization and scale pixel values to fall between 0 and 1. To make the model less prone to overfitting, we incorporate common data augmentation strategies like flipping images horizontally and vertically, slight rotations (within $\pm$15 degrees), and random brightness adjustments. These small changes help the model learn to recognize glaucoma features under a wider variety of fundus images.
\newpage
\begin{figure}[t]
    \centering
    \caption{Fundus Images}
    \includegraphics[width=\linewidth]
    {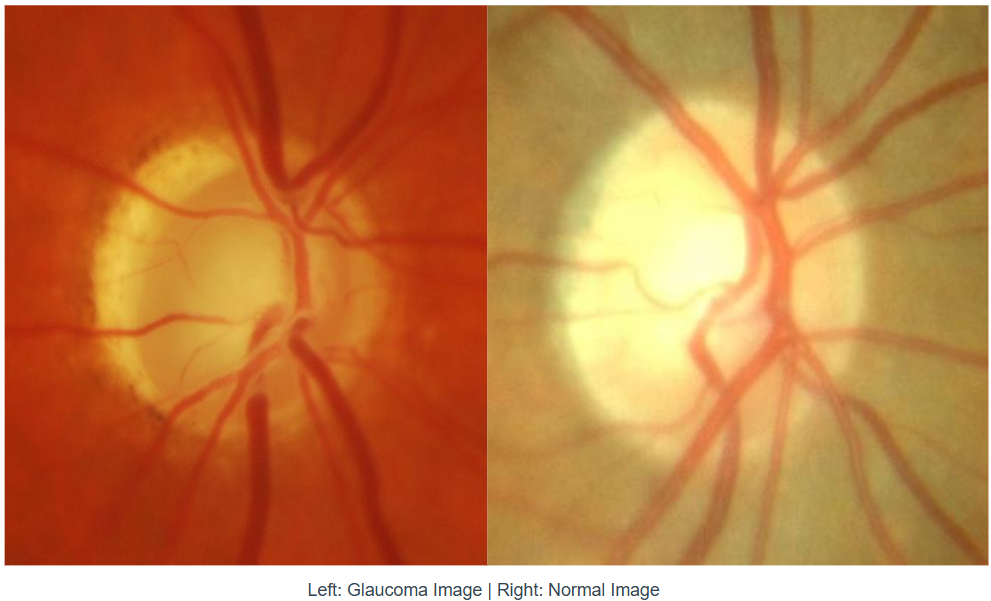}
\end{figure}
We also observe that the different datasets use slightly different labeling schemes so we would map all the labels to a simple binary format of 0 for normal and 1 for glaucomatous. This helps us to maintain uniformity during training and evaluation. A custom PyTorch Dataset class was created for managing the datasets . This class handles loading of data, preprocessing, and labeling of the images, and allowed smooth switching between preprocessing techniques and training/testing models.

\subsection{Model Architecture}
We use the exisiting EfficientNet-B0 architecture as our base model due to its balance between accuracy and computational efficiency for large datasets. The pre-trained EfficientNet model on ImageNet serves as the starting point of our design and instead of the final classification layer, we plan to add custom head of a global average pooling layer, followed by a fully connected layer (128 units, ReLU), dropout (0.4), and a final sigmoid output layer for binary classification.

This architecture is designed to capture hierarchical visual features with fewer parameters which makes it suitable for training on medical datasets without underfitting/overfitting.

\subsection{Training Strategy and Hyperparameter Tuning}
Our training process consisted of multiple phase of sequences: initially we trained our model  on ACRIMA, then fine-tuned it on ORIGA, and finally tested on the unseen RIM-ONE dataset to check generalization as discussed below:
\begin{enumerate}
    \item First training loop on the ACRIMA dataset for base feature learning of the images.
    \item Next fine-tuning on the ORIGA dataset using lower learning rates to retain the learned representations.
\end{enumerate}

We initially train on 25 epochs and gradually increase/decrease it based on the model convergence behavior. We also use the Adam optimizer with a learning rate of $1\times10^{-3}$ and binary cross-entropy loss. Optimizations like momentum and scheduler also have been applied based on model training behavior. We finally use early stopping with L2-regularization that is applied based on the validation AUC-ROC curve. Batch size is set to 32 as the datasets are not massive in size. All training is performed using PyTorch on an NVIDIA GPU environment.

\subsection{Evaluation Metrics and Reproducibility}
In order to evaluate performance, we used some important classification metrics such as—
accuracy, sensitivity, specificity and F1-score. Additionally, ROC curves were generated and compared to visualize the model’s discriminative power and error distribution. Then we compiled the final results across different experimental configurations to evaluate the effectiveness of our preprocessing strategies and the model’s ability to generalize across datasets. Also, the dataset splits into train \& test are performed in manner to preserve class distribution. Our modular pipeline using the EfficientNet-B0 architecture allows anybody to plug in new datasets or swap architectures/layers with minimal code/logic changes, enabling new features and metrics.

\section{Experiments}
We perform 2 experiments to analyze the contribution and validate the effectiveness of our proposed multi-dataset training of the CNN model using fundus images. These experiments would help us measure the model accuracy, evaluate generalization to unseen data, test different pre-processing and training strategies as proposed.

\subsection{Experiment 1: Impact of Preprocessing Techniques}
This experiment analyzes how various pre-processing methods affect the model performance. We compare models trained with minimal preprocessing of resizing and normalization against those trained with enhanced preprocessing techniques such as histogram equalization, contrast normalization, and data augmentation. The aim here is to determine how multiple preprocessing techniques can result in better accuracy and faster convergence, especially across varied datasets with differing images.
\begin{itemize}
    \item \textbf{Accuracy} helps us measure the overall performance improvement due to preprocessing.
    \item \textbf{Sensitivity \& Specificity} helps us assess how preprocessing impacts the model's ability to detect glaucoma vs normal fundus images.
    \item \textbf{AUC-ROC} curve helps us evaluate discriminative ability across the different preprocessing pipelines.
\end{itemize}
The experiment gave opposite results to our expectations as the model with minimal preprocessing achieved a higher discriminative performance with an AUC-ROC of $0.79$ that performs better than that of enhanced preprocessing, which achieved $0.73$ (see Fig 2). This shows that aggressive preprocessing can distort fine image features critical for glaucoma detection. Also, the training loss steadily decreased for the minimal pipeline, while there was significant fluctuation for the enhanced one (see Fig 3 and 4).

For the metrics calculated (see Table 1), the model with minimal preprocessing got an overall accuracy of $64.6\%$ with an AUC-ROC of $0.79$, while the enhanced one got $62.3\%$ in accuracy and $0.73$ in AUC-ROC. Furthermore, enhanced preprocessing improved sensitivity ($73.5\%$) but reduced specificity ($58.3\%$) which indicates a higher false positive rate. This gives an interesting observation where an enhanced pipeline improves recall but tends to overfit, as shown in the fluctuating test loss plots.
\begin{figure}[h]
    \centering
    \includegraphics[width=\linewidth]
    {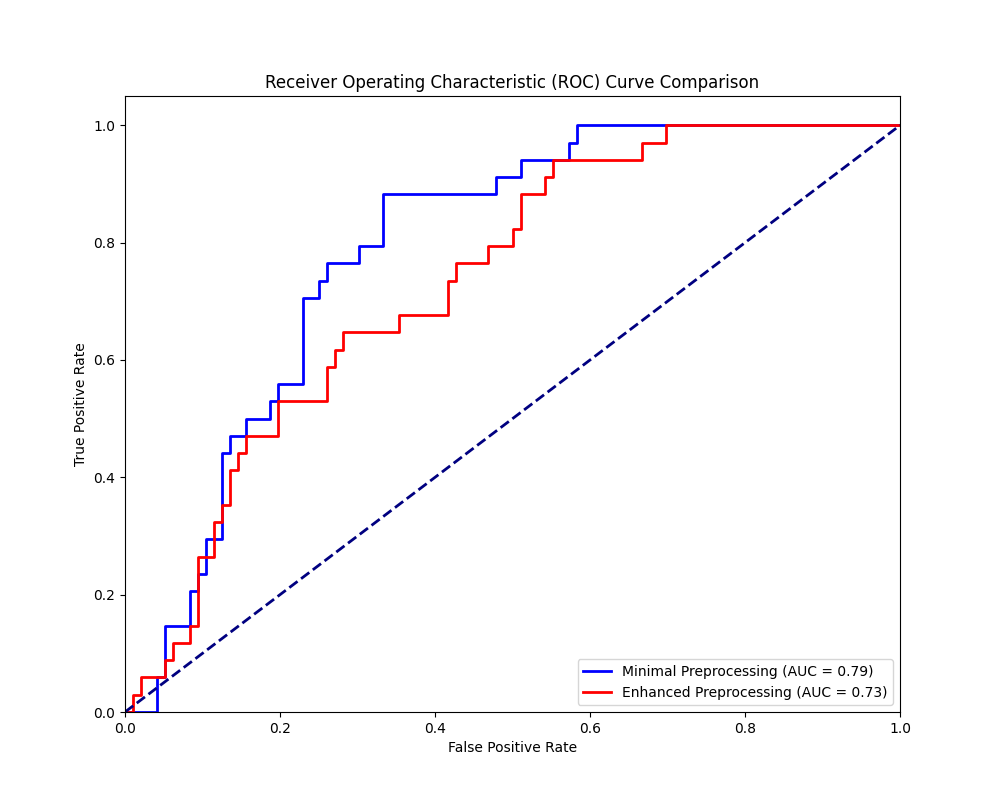}
    \caption{ROC Curves}
\end{figure}
\begin{figure}[h]
    \centering
    \includegraphics[width=\linewidth]
    {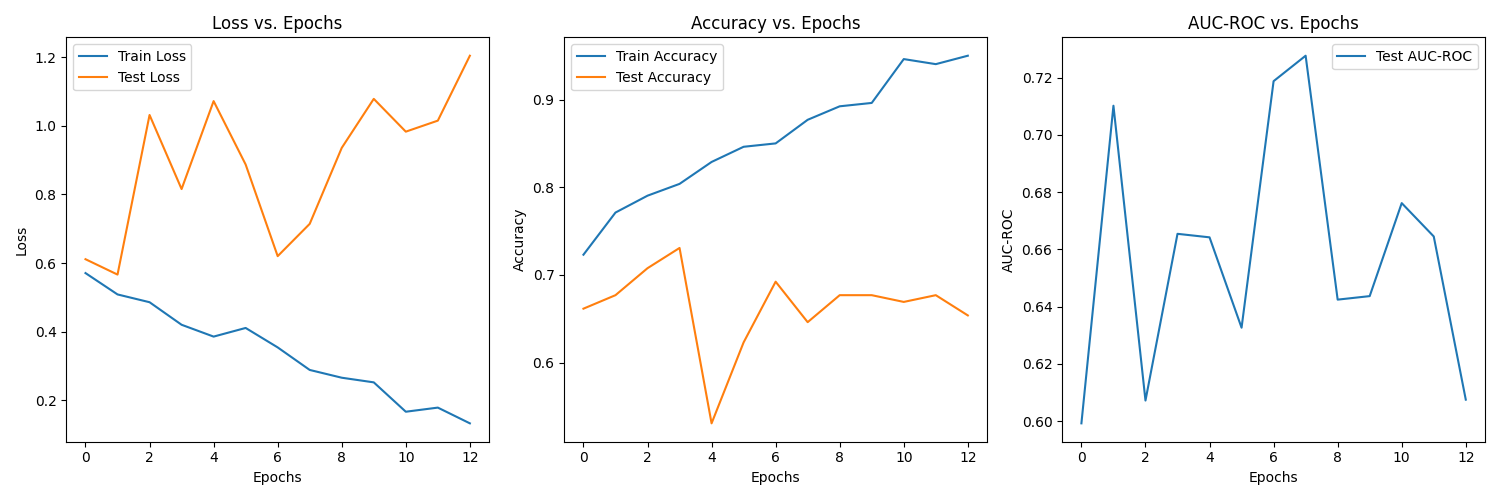}
    \caption{Minimal Preprocessing Training}
    \includegraphics[width=\linewidth]
    {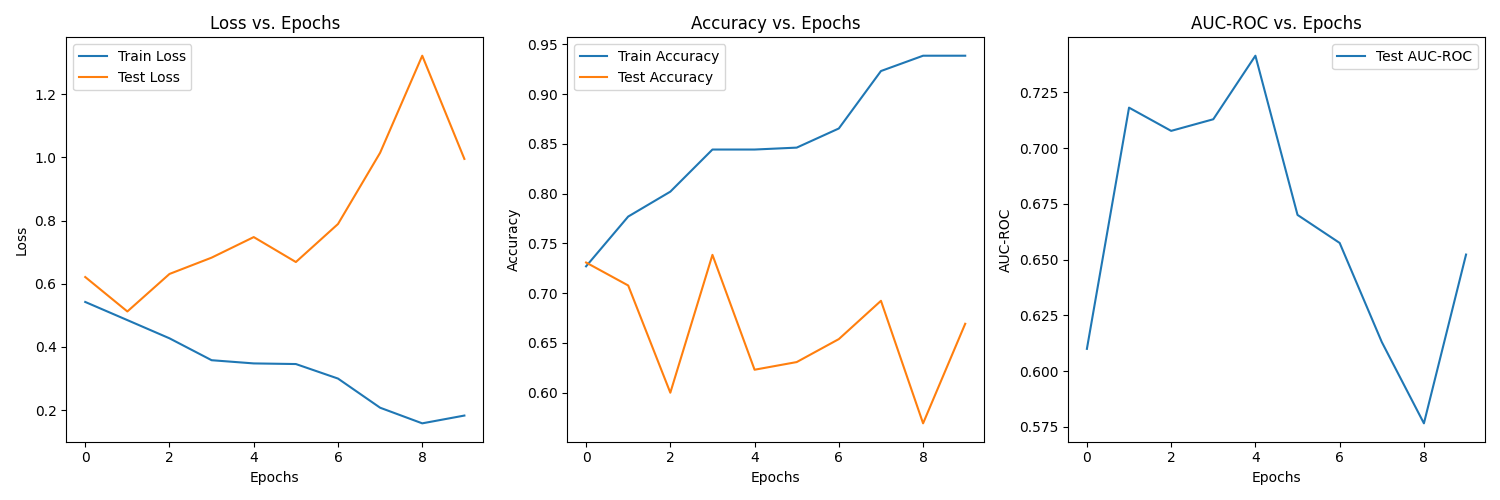}
    \caption{Enhanced Preprocessing Training}
\end{figure}
\begin{table}[h]
\centering
\caption{Performance Comparison Between Preprocessing Techniques}
\begin{tabular}{|l|c|c|c|c|}
\hline
\textbf{Experiment} & \textbf{Accuracy} & \textbf{Sensitivity} & \textbf{Specificity} & \textbf{AUC-ROC} \\
\hline
Minimal Preprocessing & 0.646 & 0.676 & 0.635 & 0.789\\
Enhanced Preprocessing & 0.623 & 0.735 & 0.583 & 0.728\\
\hline
\end{tabular}
\label{tab:preprocessing_results}
\end{table}

\newpage
\subsection{Experiment 2: Generalization to Unseen Dataset}
This experiment evaluates how well the trained model generalizes to an unseen dataset not included in training. We train the models on ACRIMA and ORIGA datasets and test on the RIM-ONE dataset. This simulates real-world conditions where the model may be deployed on new data distributions.
\begin{itemize}
    \item \textbf{AUC-ROC} curve helps us assess how well the model separates classes in the unseen dataset.
    \item \textbf{F1-Score} helps us measure the balance between precision and recall metrics of model performance, giving crucial information when the datasets are imbalanced.
\end{itemize}
As per expectations, the test data performance showed a near-perfect score, AUC $=0.99$ for ACRIMA, while ORIGA and RIM-ONE scored $0.73$ and $0.88$, respectively (see Figure 4). This confirmed our hypothesis that the performance is strongly dependent on the dataset characteristics, with the ACRIMA dataset being highly optimal. The comparison of evaluating metrics on model performance across the datasets shows a declining trend from ACRIMA → RIM-ONE → ORIGA (see Fig 6).

The training history for each dataset (see Figures 7–9) confirms our hypothesis as it shows overfitting trends on ORIGA and RIM-ONE, where test losses increase over epochs while train losses decrease. On training the model upon ACRIMA and ORIGA while evaluating test generalization to the RIM-ONE dataset (see Table 2), the model achieved an AUC-ROC of $0.88$ and an F1-score of $0.60$, which confirms a decent discriminative power even on unseen data. But, as observed the sensitivity dropped to $48.2\%$, which indicates issues in detecting glaucoma in different imaging styles. Upon comparing, the model performs best on ACRIMA (AUC = 0.99, F1 = 0.99), followed by ORIGA (AUC = 0.73, F1 = 0.51) thus highlighting the benefit of a multi-dataset training pipeline while stressing the need for adaptive tuning.
\begin{table}[h]
\centering
\caption{Dataset-wise Evaluation of the Trained Model}
\begin{tabular}{|l|c|c|c|c|c|}
\hline
\textbf{Dataset} & \textbf{Accuracy} & \textbf{Sensitivity} & \textbf{Specificity} & \textbf{F1-Score} & \textbf{AUC-ROC} \\
\hline
ACRIMA & 0.993 & 0.987 & 1.000 & 0.994 & 0.999 \\
ORIGA & 0.677 & 0.647 & 0.688 & 0.512 & 0.731 \\
RIM-ONE & 0.799 & 0.482 & 0.949 & 0.607 & 0.878 \\
\hline
\end{tabular}
\end{table}

\newpage
\begin{figure}[H]
    \centering
    \includegraphics[width=\linewidth]
    {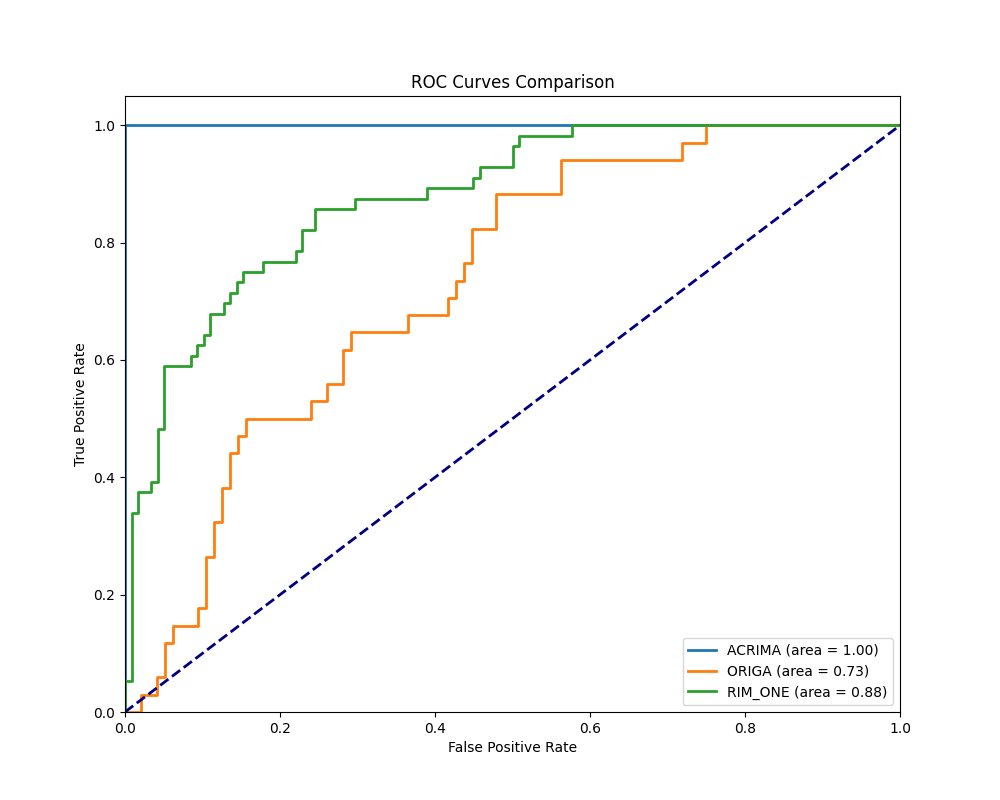}
    \caption{ROC Curves}
    \includegraphics[width=\linewidth]
    {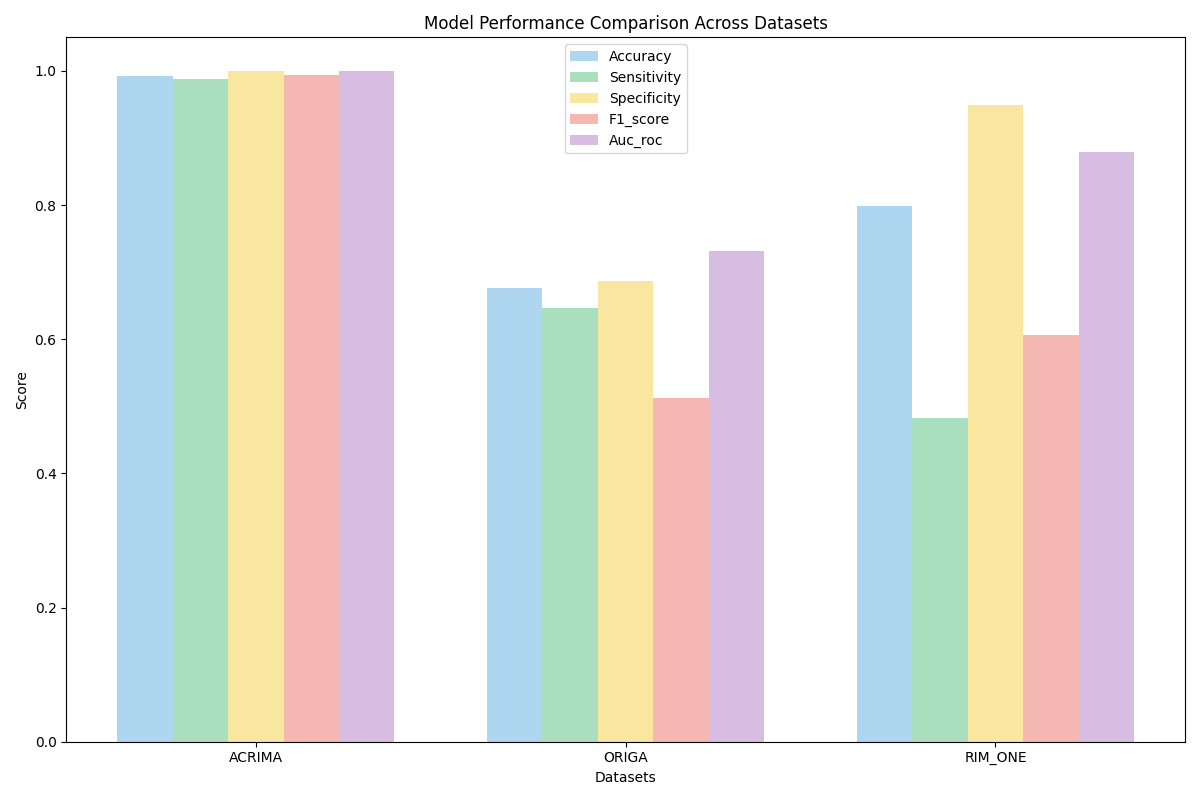}
    \caption{Model Performance Comparison}
\end{figure}

\begin{figure}[H]
    \centering
    \includegraphics[width=\linewidth]
    {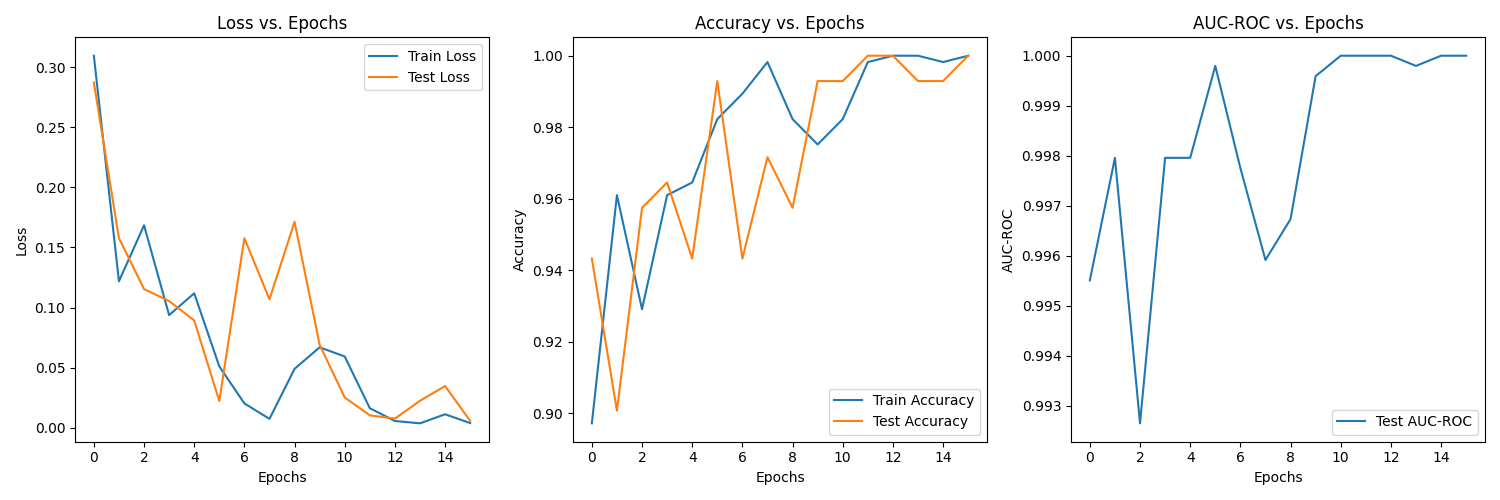}
    \caption{ACRIMA Training}
    \includegraphics[width=\linewidth]
    {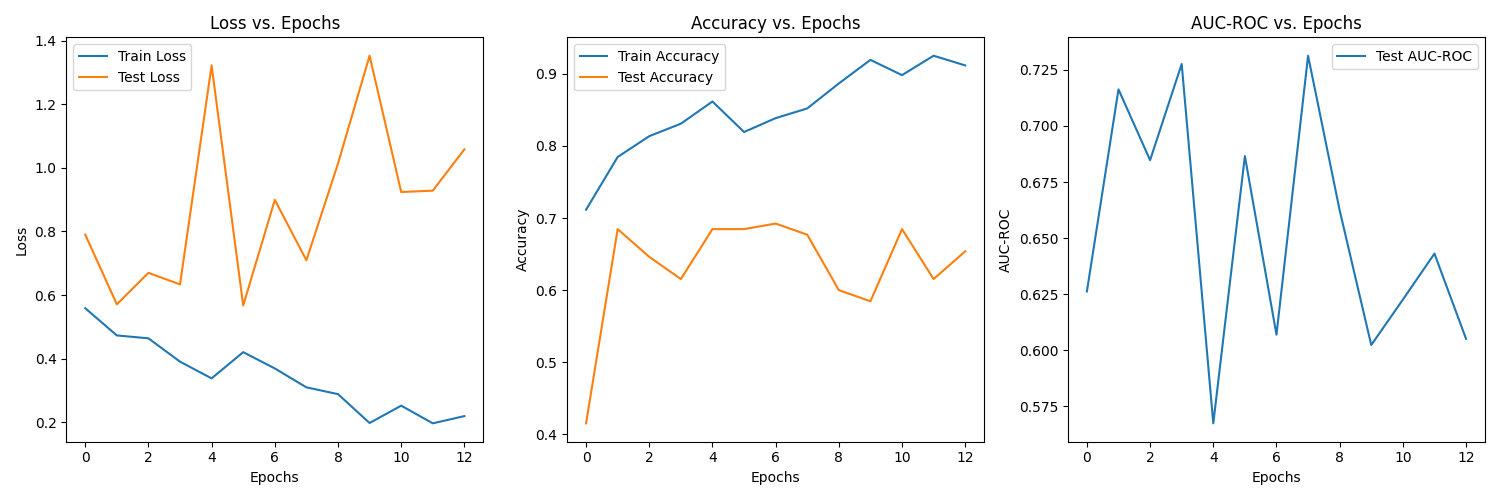}
    \caption{ORIGA Training}
    \includegraphics[width=\linewidth]
    {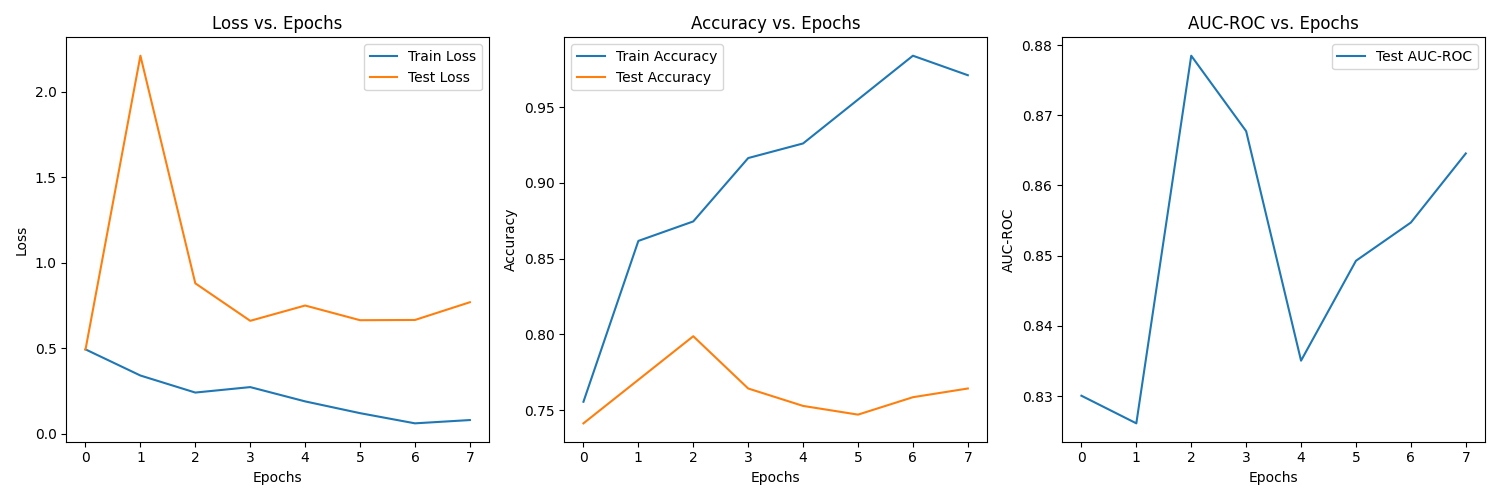}
    \caption{RIM-ONE Training}
\end{figure}

\subsection{Limitations and Next Steps}\
The performed experiments clearly show the benefit of a multi-dataset training pipeline and the interesting outcome in case of enhanced preprocessing of image datasets. Some experiments that still need to be considered are, first our existing model does not fully explore domain adaptation or fine-tuning strategies that could improve generalization to underrepresented datasets like ORIGA and, second the impact of individual preprocessing operations like histogram equalization vs. data augmentation for the proposed model is missing.

Thus, this baseline CNN architecture can be further extended to new inclusions like systematic analysis of preprocessing steps to understand their individual contributions, incorporating domain adaptation techniques such as adversarial training or feature alignment, expanding the dataset to include imaging variations that ensures broader generalization and evaluating model fine-tuning to ensure confidence in predictions. Together, these steps can help us make our model more robust and clinically deployable in real-world glaucoma screening situations.
\section{Conclusion}

In this project, we develop a deep learning-based Efficient-Net architecture for early glaucoma detection using fundus images of the eye. Prior models relied and trained on single datasets whereas our methodology uses multiple publicly available datasets i.e. ACRIMA, ORIGA and RIM-ONE to train a CNN-model that is more generalizable and accurate. We follow a sequential fine-tuning strategy on the multiple datasets and different dataset preprocessing strategies that aims to address the limitations of dataset-specific bias and improve real-world clinical performance of the model. This work on a baseline deep-learning model of EfficientNet-B0 lays the groundwork for a scalable, accessible, and accurate glaucoma screening tool for preventive detection that can be extended across to diverse patients and fundus images.

This glaucoma detection model is compatible with early detection and makes this technology more reachable; however, some moral issues come along with it, which is important for us to consider. We should not only focus on any of the specific groups, but our model needs to perform well across the patients of different groups and eye conditions to ensure fairness. We should also consider accountability as well, since being a medical equipment, it should only act as a supplement to a doctor's expertise rather than replacing them. Transparency is essential because doctors and patients should understand how the system comes to its decisions to build trust among its users. Even when using publicly available images, it is important to maintain patient confidentiality and ensure that their data is protected by data privacy. Finally, we must be careful of the broader social impact of introducing such AI-driven tools in healthcare, making sure they help closely, rather than widening the gaps in accessibility and quality of care.

\bibliographystyle{plain}
\bibliography{egbib}       
\end{document}